\documentclass[conference]{IEEEtran}
\usepackage{times}

\usepackage[numbers]{natbib}
\usepackage{multicol}
\usepackage[bookmarks=true]{hyperref}
\usepackage{subcaption}
\usepackage{hyperref}
\usepackage{url}
\usepackage{multirow}
\usepackage{array} 
\usepackage{booktabs}
\usepackage{wrapfig}
\usepackage{graphicx}
\usepackage{tcolorbox}
\usepackage{makecell}

\usepackage{tikz} 

\makeatletter
\def\blfootnote{\xdef\@thefnmark{}\@footnotetext}
\makeatother

\tcbuselibrary{listings, breakable, skins}
\newcommand{\projname}[0]{EgoActor}
\newcommand{\taskname}[0]{EgoActing}

\newcommand{\ie}{\emph{i.e., }}
\newcommand{\eg}{\emph{e.g., }}

\begin{document}

\title{\projname{}: Grounding Task Planning into Spatial-aware Egocentric Actions for Humanoid Robots via Visual-Language Models}

\author{Yu Bai, MingMing Yu, Chaojie Li, Ziyi Bai, Xinlong Wang$^{\dagger}$, Börje F. Karlsson$^{\dagger}$\\
\normalsize{Beijing Academy of Artificial Intelligence}}



%


\twocolumn[{%
\renewcommand\twocolumn[1][]{#1}%
\maketitle
\vspace{-0.5cm}
\begin{center}
    \centering
    \captionsetup{type=figure}
    \includegraphics[width=\linewidth]{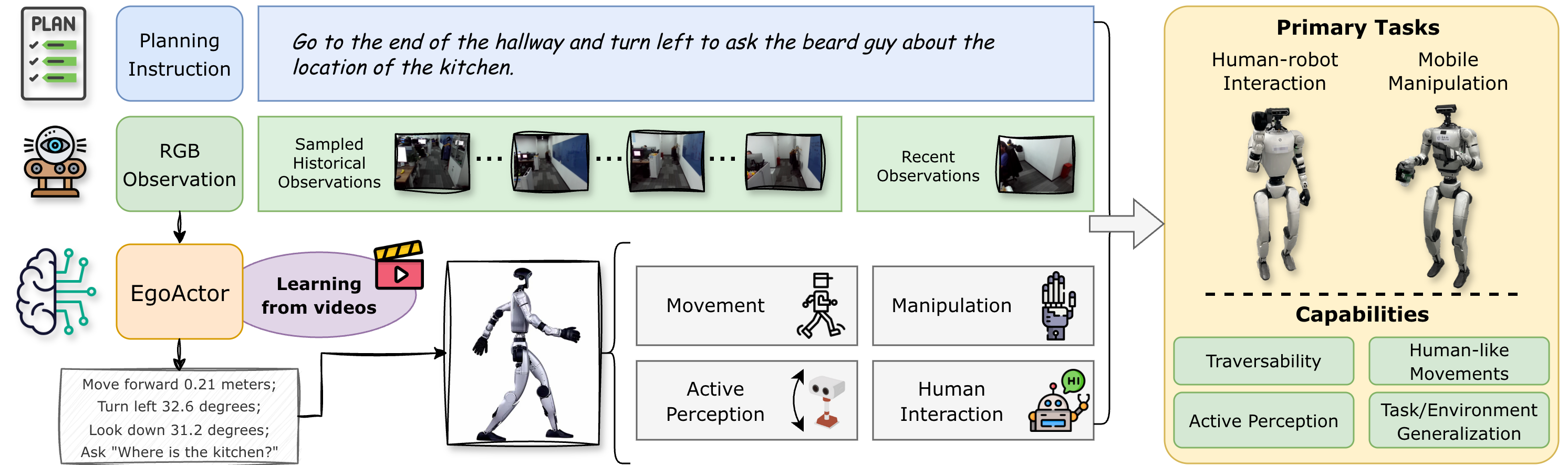}
    \captionof{figure}{Overview of \projname{}, which can control a humanoid robot by jointly predicting movement, active perception, manipulation, and human interaction actions to achieve coordinated and precise execution, enabling humanoid robots to conduct long-horizon multi-step task instructions described in natural language.}
    \label{fig:intro}
\end{center}
}]

\begin{tikzpicture}[remember picture, overlay]
    \node[anchor=north west, yshift=-0.2in, xshift=0.7in] at (current page.north west) {
        \includegraphics[width=3cm]{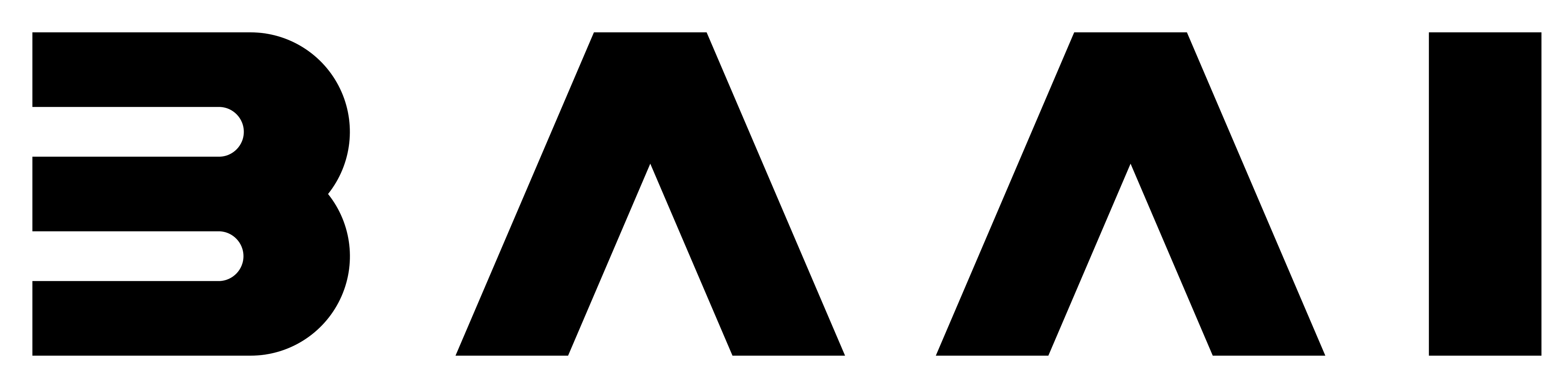}
    };
    
    \draw[darkgray] (current page.north west) ++(0.75in, -0.6in) -- ++(\columnwidth * 2.0, 0);
\end{tikzpicture}

\begin{abstract}
Deploying humanoid robots in real-world settings is fundamentally challenging, as it demands tight integration of perception, locomotion, and manipulation under partial-information observations and dynamically changing environments. As well as transitioning robustly between sub-tasks of different types.
Towards addressing these challenges, we propose a novel task — \taskname{}, which requires directly grounding high-level instructions into various, precise, spatially aware humanoid actions.
We further instantiate this task by introducing \projname{}, a unified and scalable vision-language model (VLM) that can predict locomotion primitives (\eg walk, turn, move sideways, change height), head movements, manipulation commands, and human-robot interactions to coordinate perception and execution in real-time.
We leverage broad supervision over egocentric RGB-only data from real-world demonstrations, spatial reasoning question–answering, and simulated environment demonstrations, enabling \projname{} to make robust, context-aware decisions and perform fluent action inference (under 1s) with both 8B and 4B parameter models.
Extensive evaluations in both simulated and real-world environments demonstrate that \projname{} effectively bridges abstract task planning and concrete motor execution, while generalizing across diverse tasks and unseen environments\footnote{We open-source our code, models, datasets, and benchmarks to facilitate future research: \url{https://baai-agents.github.io/EgoActor/}.}.

\end{abstract}

\IEEEpeerreviewmaketitle
\blfootnote{$\dagger$ Corresponding authors. \{borje, xlwang\_1\}@baai.ac.cn}

\section{Introduction}

Deploying humanoid robots in real-world environments and tasks is fundamentally challenging~\citep{gr00tn1_2025, zhang2025flam0, yuan2025being000}.
These challenges mainly arise from the inherent instability of humanoid platforms and the complexity of partial-information real-world tasks.
Humanoid robots are inherently unstable compared to wheeled platforms and are highly sensitive to issues in timing, precision, and obstacle handling. Even minor control inaccuracies can disrupt balance or lead to unsafe behaviors, especially in dynamic, cluttered, and previously unseen environments. While recent advances in low-level control—such as whole-body locomotion for balance~\citep{cheng2024expressive, liao2025beyondmimic, luo2025sonic}, and dexterous hand-based manipulation~\citep{intelligence202500_0005000, jiang2025dexmimicgen, huang2025human}—have substantially improved motor execution, approaches for these capabilities usually focus only on that task type only, and still require precise coordination based on spatial and visual understanding, thus remaining fragile under real-world uncertainty. 

Beyond basic locomotion and manipulation, embodied robot systems also require fluid transitions between different actions and action types, often executing them in combination. Real-world tasks rarely involve actions in isolation~\citep{yuan2025being000}; instead, they require the coordinated use of movement, head orientation, hand manipulation, human-robot interactions (\eg gestures and utterances, etc.), and full-body control in contextually appropriate sequences. For example, a robot may need to halt its walking, tilt its head to perceive a target, extend its arm to grasp an object, and then resume locomotion—all as part of a single coherent task~\citep{wang2025instruction0augmented}. Achieving such coordination demands not only reliable motor control, but also advanced spatial understanding and reasoning, enabling a model to infer how actions interact with the environment
and to determine when and how to effectively coordinate multiple skills.

To embody the above challenges, we introduce \taskname{}, representing scenarios in which a humanoid robot must transform an actionable instruction into appropriate situated action sequences, based on egocentric observations, action history, and its available skills. Moreover, to instantiate this task, while allowing naturally scalable data collection and future expansion, we develop \textit{\textbf{\projname{}}}, a vision–language model (VLM)~\citep{bordes2024introduction} that grounds high-level instructions into low-level, executable humanoid actions. We leverage the reasoning capabilities of VLMs, while enhancing their spatial understanding to directly predict a wide range of low-level actions formulated in language, including \textit{movement, active perception, manipulation, and human-interaction} (as shown in Fig.~\ref{fig:intro}). For movement, \projname{} outputs precise locomotion primitives such as moving forward or strafing by specified distances, turning by specified angles, and performing postural adjustments like standing or crouching to support manipulation at different heights. For active perception, it predicts head orientation actions to facilitate exploration, target localization, and dynamic obstacle handling. For manipulation, it determines when and where to initiate hand or arm actions for coordinated object interaction. Additionally, it can produce human interaction actions that enable information seeking, communication, and collaboration with humans or other robots through gestures or spoken language.

To equip the model with these capabilities, we train \projname{} on a diverse mixture of real-world video demonstrations, spatial reasoning trajectories, action-timing annotations, and virtual environment examples; minimizing human annotation requirements. This broad supervision enables fully functional 8B and 4B models with sub-second inference latency, supporting real-time interaction and control on humanoid platforms. 
We evaluate the proposed framework across a wide range of settings, including human–robot interaction\footnote{When referring to the task setting, we use the term \emph{human–robot interaction}. When referring to a candidate skill or action from the robot’s perspective, we term it \emph{human-interaction}.}, mobile manipulation, and traversability—defined as the robot’s ability to safely move through narrow spaces commonly encountered in daily environments without colliding with surrounding obstacles—in real-world and simulated environments. These experiments demonstrate task- and environment-level generalization, highlighting \projname{}’s ability to operate under diverse and unseen conditions. In addition, qualitative case studies showcase behaviors such as active perception and human-like movement patterns. 

We summarize our contributions as follows:

\begin{itemize}
    \item We introduce \taskname{}, a new task formulation that requires models with strong spatial understanding to directly transform language instructions into executable action sequences from egocentric observations, emphasizing the challenges of real-world humanoid deployment.
    \item We propose \projname{}, a vision-centric model that fully leverages humanoid capabilities by unifying movement, perception, manipulation, and human interaction.
    \item We validate \projname{} through extensive real-world and simulated experiments, and release deployable open-source code and models, along with datasets and evaluation protocols to facilitate reproducibility and future research.
\end{itemize}

\section{Related Work}
\subsection{VLM-based Embodied Agents}
Vision-and-Language-Model (VLM) based embodied agents aim to ground natural language instructions into executable actions~\citep{NEURIPS2023_efb2072a,wu2023embodied, song2023llm}. A representative example is SayCan \citep{ahn2022can}, which decomposes language instructions into executable skills using learned affordances; however, such approaches typically rely on predefined skill libraries and are therefore less suitable for humanoid robots with complex and diverse embodiments. Recent surveys \citep{szot2024multimodalllmsgeneralistembodied, fung2025embodiedaiagentsmodeling} review progress in multimodal embodied agents and identify key challenges in grounding, long-horizon reasoning, and real-world deployment. Building on this line of research, several works explore LLM-based embodied agents in simulated or structured environments, including cooperative multi-agent systems \citep{guo2024embodied}, standardized benchmarking interfaces \citep{li2024embodied}, offline reinforcement learning with LLM-generated rewards \citep{lee-etal-2024-llm}, language-supervised policy learning \citep{yang2024embodied}, and LLM-driven environment generation \citep{zala2024envgen}. Other studies target specific capabilities, such as zero-shot object navigation~\citep{dorbala2023can} and visual perception in open-world~\citep{zheng2023steve}.

 In contrast, \projname{} focuses on humanoid robots and directly predicts egocentric, low-level executable actions such as locomotion and head movement, bridging textual task description and low-level motor control.

 \subsection{Mobile-Manipulation}
Mobile manipulation has been extensively explored across simulated and real-world frameworks. Early systems such as SayCan~\citep{ahn2022can} combine language models with affordance-based planners to ground high-level instructions into sequential robot skills. 
Simulation platforms ~\citep{ehsani2021manipulathor,deitke2022,szot2021habitat,yenamandrahomerobot,li2024behavior01k0} provide embodied environments for evaluating navigation and manipulation jointly. 
Recent efforts~\citep{yang2024harmonic, melnik2023uniteam, gumaniskill2} further expand large-scale benchmarks, emphasizing diverse object interactions, long-horizon tasks, and more realistic robot embodiments.


In contrast to previous work,
\taskname{} and \projname{} unify decision-making at an egocentric level, jointly reasoning over locomotion, posture, spatial perception, manipulation, and human interaction within a single VLM-based action predictor. This integrated approach enables smooth movement–manipulation transitions and robust adaptation to unseen layouts and diverse natural-language instructions.

\subsection{Visual-Language Navigation}
\paragraph{Vision-and-Language Navigation} Vision-and-Language Navigation (VLN) has been extensively studied as a core embodied AI problem~\citep{gu2022vision, zhangvision}.
Classical VLN methods predominantly focus on mapping language instructions to navigational trajectories without requiring physical interaction with the environment. Representative works include R2R~\citep{anderson2018vision}, Touchdown~\citep{chen2019touchdown}, and sub-instruction-aware navigation~\citep{hong2020sub}, all of which aim to follow language-guided routes in static or semi-structured environments.
More recent approaches extend VLN with stronger visual–language models and improved spatial grounding. VLN-R1~\citep{qi2025vln0r10}, NaVid~\citep{zhang2024navid0}, Uni-NaVid~\citep{zhang2025uni0navid0}, and Navila~\citep{cheng2024navila0} leverage large multimodal encoders and unified navigation architectures to improve generalization, robustness, and long-horizon reasoning. 

\paragraph{Object-goal Navigation} Object-goal navigation requires an agent to navigate to a specified object category using only its onboard perception~\citep{sun2024survey, yu2025cnav}. 
\citet{chaplot2020object} propose a modular semantic-mapping system that leverages object-arrangement priors for efficient navigation in unseen environments. \citet{qi2020object} introduce OAAM, which separately encodes object and action descriptions to improve language-vision alignment in VLN. \citet{cao2024cognav0} use cognitive-state modeling with a dynamic map and an LLM to guide navigation via map state reasoning, enhancing success in both simulation and real-world settings.

In contrast, \projname{} extends far beyond navigation: it coordinates whole-body humanoid behaviors by jointly reasoning over locomotion, posture control, active perception, manipulation, and human-interaction behaviors. This enables the model to ground high-level instructions into actionable, egocentric motor sequences suitable for real-world humanoid control in dynamic environments.

\section{Framework Design}   
\subsection{Task Definition}
\label{sec:task_definition}
\begin{figure*}[t]
\centering
\includegraphics[width=0.95\linewidth]{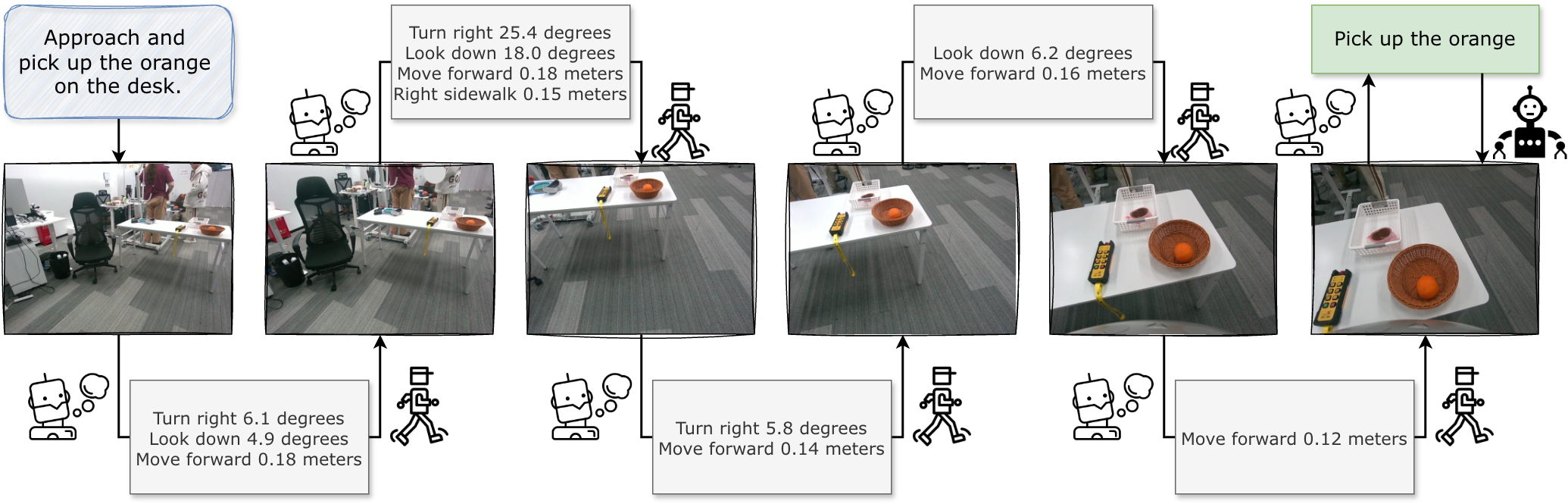} 
\caption{Visualization of \projname{}'s working procedure for a given task: ``Approach and pick up the orange on the desk''. The grey blocks represent structured language actions (SLAs) and the green blocks represent natural language actions (NLAs).}
\label{fig:procedure}
\end{figure*}

In this work, we introduce \taskname{}, a task that receives a direct and actionable instruction and predicts the next concrete actions a humanoid robot should perform. We assume the robot is equipped with a set of whole-body control and manipulation policies. The task could be represented as follows: 
\begin{equation}
a_t = \arg\max_{a \in \mathcal{A}} \; P (a \mid I,\; O_{1:t},\; a_{1:t-1},\; \Pi)
\label{eq:egoacting_prob}
\end{equation}
where $I$ is the (natural-language) task instruction, $O_{1:t}$ denotes the history of egocentric observations (\ie RGB images in our setup), $a_{1:t-1}$ is the past action history, and $\Pi$ denotes the set of available low-level whole-body and manipulation policies. An example of the proposed \projname{} conducting an \taskname{}-type task is shown in Fig.~\ref{fig:procedure}.

\textbf{Instruction.} We define an instruction as a high-level yet explicit specification of the intended task, describing the required movements and goals without prescribing low-level motor details. The instruction captures general motion primitives—such as passing through a corridor, turning at specific locations, or stopping to manipulate objects and interact at designated positions—thereby providing clear spatial and temporal guidance. When a target is involved, the target object is specified in an unambiguous and identifiable manner. 
This design discourages the model from relying on guesswork or producing hallucinated predictions, and instead promotes grounded reasoning and reliable action execution.


Especially, we include the following basic capability categories as source of candidate actions in the current task (specific examples shown in Appendix~\ref{app:egoacting_sample}):

\paragraph{Active perception} We introduce active perception skills that allow the humanoid robot to better explore its surroundings, localize target objects, and dynamically respond to newly appearing obstacles along its path~\citep{sen2024learning, wang2023active}.

\paragraph{Manipulation} We include manipulation as an essential action type as it plays a crucial role in humanoid robotics by allowing coordinated control of the hands and arms to perform precise interactions with objects and tools in unstructured environments~\citep{luo2025being0h00}.

\paragraph{Human-interaction} We also include human-interaction actions, allowing the robot to seek new information, engage in communication, and cooperate with humans or other humanoid robots by requiring items from others.

\paragraph{Movement} For movement actions, we design a set of basic locomotion skills (summarized in Appendix~\ref{app:supported_skills}). Beyond conventional navigation tasks, \taskname{} also incorporates lateral movement actions. This capability enhances obstacle avoidance and enables more precise alignment with target objects. Meanwhile, to support manipulation tasks across varying heights and spatial constraints, we include postural adjustment actions such as standing up and crouching down. These skills help the robot adapt to tasks that require interacting with objects at different elevations or avoiding overhead obstacles.



\subsection{Language-based Actions}

We design the framework to represent robot behaviors as textual actions, combining structured language actions for precise movement and perception with natural language actions for manipulation and human interaction. 

\paragraph{Structured language actions (SLAs)}
For movement and active perception, we adopt a set of structured language actions that describe spatial motion in a well structured and interpretable format. Each action is expressed in a concise natural-language-like template specifying the action type, direction, and magnitude, such as:
\textit{Turn left 30.5 degrees} or \textit{Look up 10.2 degrees}, details shown in Appendix~\ref{app:supported_skills}.

These structured language actions encompass horizontal and vertical rotations (yaw and pitch), linear translations along the forward–backward and lateral axes, and vertical adjustments along the z-axis. Thresholds are applied to filter out negligible movements and reduce noise. The purpose of these actions is to enable the model to interpret spatial relationships from RGB observations and to position the robot appropriately for executing subsequent task-specific (natural language) actions.

\begin{figure}[t]
\centering
\includegraphics[width=0.95\linewidth]{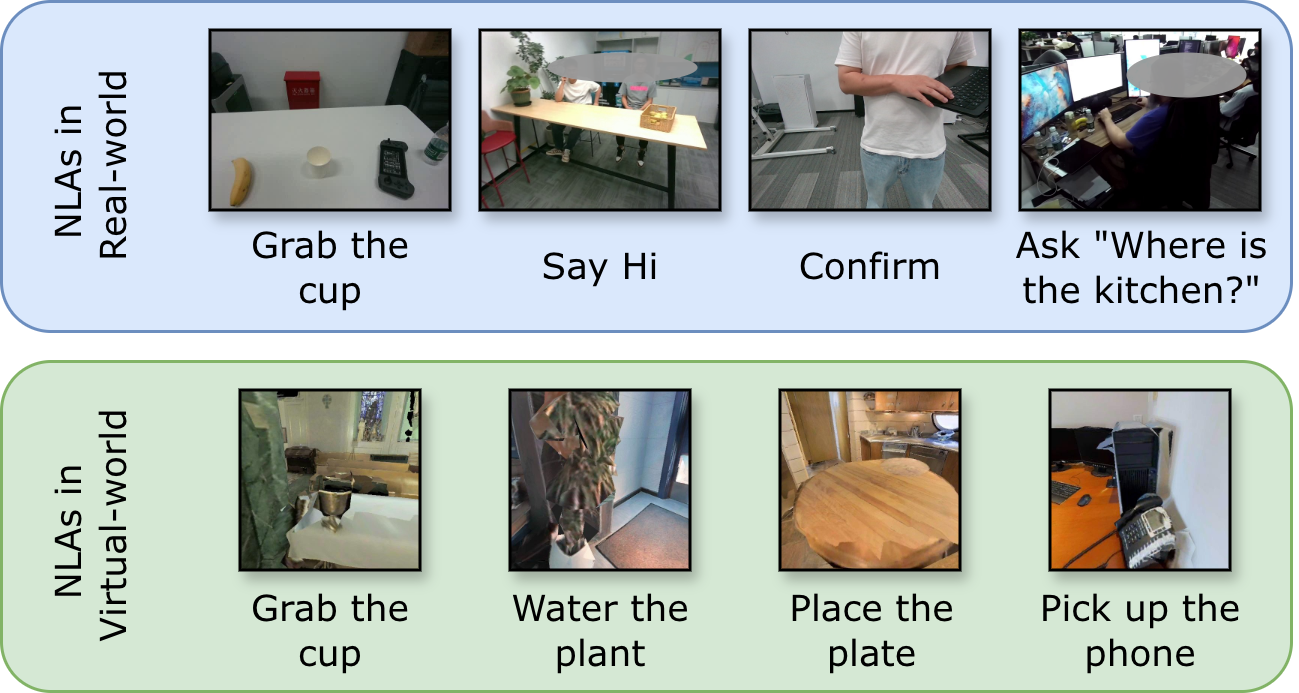} 
\caption{Example natural language actions (NLA) in \taskname{}. \projname{} is trained to predict the corresponding actions based on obtained RGB observations.}
\label{fig:ending_action}
\end{figure}

\paragraph{Natural language actions (NLAs)}
For manipulation and human-interaction actions, we do not restrict the system to a fixed set of skills. Instead, we employ natural language to represent these actions, with examples shown in Fig.~\ref{fig:ending_action}. This design provides the following key advantages:
\begin{itemize}

    \item Generalization beyond predefined skill or code primitives, enabling the interpretation of novel instructions and the production of previously unseen actions.
    \item Effective reuse of low-level vision–language–action models for manipulation by precise pre-positioning and providing context-aware language commands for complex, open-ended interactions.
    \item Transformation of task intentions into natural language actions. For example, for the task ``Search and approach the woman and ask her to show you the way to a meeting room'', the model can output ``Ask \textit{Could you please guide me to a meeting room?}''. More details are provided in Appendix~\ref{app:human_interaction_case}.
\end{itemize}

We also include ``Stop and no action'' as an NLA, to mean the task is done and the robot should wait for a new instruction.

\paragraph{Discussion} We note that the primary role of structured language actions is to navigate and position the robot to enable subsequent natural language actions, which requires the model to possess strong spatial understanding ability. In most cases, each data sample consists of one sequence of structured language actions and one subsequent natural language action. We also include interleaved-style data from the EgoTaskQA dataset~\citep{DBLP:conf/nips/JiaLZH22}, which features multiple alternating sequences of structured and natural language actions. In the former case, we treat the natural language action as an implicit stop signal when it appears at the end of the action sequence; otherwise, and for the interleaved-style data, we explicitly append a “Stop and no action” at the end of the sequence.

\paragraph{Action Parsing} During real-world experiments, we use a simple parser to extract parameters from SLAs, converting them into velocity/angle commands for the robot to conduct. 
For NLAs, execution is routed by keyword triggers. Actions with speech-related keywords (\eg ``Speak'', ``Ask'') are converted to audio via text-to-speech models, predefined interaction keywords (\eg ``Say Hi'', ``Shake Hands'') invoke preset motions, and all remaining actions are treated as manipulation commands, forwarded to pre-trained VLA models.

\section{Training Recipe}
In this section, we describe the model design for addressing the \taskname{}-type tasks defined in this work. We deliberately adopt a general vision–language model (VLM) architecture to demonstrate that (1) our approach does not rely on specialized architectural modifications, and (2) the model can naturally benefit from additional easily-acquirable video training data, enabling straightforward scaling in future work. 
\subsection{Model Structure and Training Setup}
We use Qwen3-VL~\citep{qwen3technicalreport}, a vision–language model built on a transformer-based architecture~\citep{vaswani2017attention} with dynamic resolution support, as our base model.
Following \citet{schulman2025lora}, we apply LoRA~\citep{hu2021lora0} to finetune all linear layers of the model with a learning rate of 3e-4. The model is trained with one epoch of randomly mixed data from all sources (see Section~\ref{sec:data-acquisition}) on 16 A100 40GB GPUs. We train both 4B and 8B variants of the model to accommodate different use cases, as we observe a trade-off between inference speed and performance across model sizes.

\subsection{Data Format} 

Generally, we format our data as illustrated in Fig.~\ref{fig:intro}, with the detailed template provided in Appendix~\ref{app:data_format}. For each task, we uniformly sample 10 historical observations from all previous images and use the most recent 3 observation-action pairs as key anchors for the model to predict the next action. To save computational resources while increasing the model’s adaptability to different hardware conditions, the recent observation images are processed at 480p resolution, while all sampled historical observation images are processed at 240p resolution. We note that this setup can be extended if practitioners have access to more computational resources and higher-quality cameras.

We manually annotate demonstrations collected from both virtual and real-world environments (see Section~\ref{sec:data-acquisition}). For each example, we first annotate a concise textual description of the pre-shot video trajectory, which can be captured using standard RGB cameras (\eg mobile phones). As mentioned in Section~\ref{sec:task_definition}, these descriptions are intentionally kept minimal yet precise, ensuring that the target object and the movement route are clearly specified and unambiguous to discourage the model from relying on guessing or producing hallucinated predictions. In addition, we append a final natural language action to each trajectory, enabling the model to explicitly associate movement sequences with subsequent manipulation or human-interaction actions. Owing to this lightweight and easily scalable annotation pipeline, our approach naturally supports large-scale data collection and is well-suited to further performance improvements through increased training data. Note that we apply additional pre-processing to the EgoTaskQA dataset, as each video trace contains multiple natural language actions (see Appendix~\ref{app:egotaskqa}).

Our approach suggests that using multiple observation-action pairs offers several advantages that previous work lacks. First, it enables more efficient training by allowing the model to learn to predict multiple actions within a single training sample. 
Second, it provides a richer context for decision-making—for example, the model can learn to have the robot turn left after previously turning right to avoid obstacles.

\subsection{Data Acquisition}
\label{sec:data-acquisition}
We utilize a diverse collection of multimodal datasets to train our model. 
For all movement data, we follow \citet{cheng2024navila0} and extract step-by-step movement actions by estimating camera poses using MASt3R~\citep{leroy2024grounding}, identifying actions at 1.5-second intervals.
For all \taskname{} sub-datasets, we augment them by oversampling samples with turning actions and natural language actions to balance the distribution across different action types.
We summarize all the data sources as follows:

\paragraph{Internet video data}
We adopt the EgoTaskQA dataset~\citep{DBLP:conf/nips/JiaLZH22} as our primary source of internet-scale egocentric videos, and supplement it with 130 additional internet-collected egocentric videos.
After processing, we produce 160,000 \taskname{} training samples from EgoTaskQA and 7,111 additional samples from the additional collected videos. Additional details on dataset preprocessing and sample construction are provided in Appendix~\ref{app:egotaskqa}.

\paragraph{Local environment data (\taskname{})}
We recorded 398 egocentric videos in local environments, yielding a total of 150,214 \taskname{} training samples. These recordings capture environmental variability due to frequent layout changes in the data collection areas. 

\paragraph{Virtual environment data (Navigation)}
To incorporate controlled spatial navigation supervision, we sampled approximately 3\% of the VLN-CE (Room-to-Room) training set~\citep{krantz_vlnce_2020}, resulting in 60,000 training samples. This subset provides diverse indoor layouts and structured navigation instructions. Detailed processing is provided in Appendix~\ref{app:virtual_env_processing}.

\paragraph{Virtual environment data (\taskname{})}
We manually collected and annotated 714 \taskname{}-style trajectories from the Habitat-Sim simulator~\citep{szot2021habitat} using scenes from the Room-to-Room dataset~\citep{anderson2018vision}. Following the VLN-CE scene-split protocol, we partition the data into 509 training trajectories and 205 validation trajectories from unseen environments. This training set resulted in 76,821 \taskname{} samples. 

\paragraph{Spatial reasoning data (MindCube)}
To strengthen the model’s spatial reasoning capabilities, we incorporate samples from the MindCube dataset~\citep{yin2025spatial}. We randomly sample 50\% of its training set, leading to 44,160 spatial reasoning samples.

\paragraph{Visual-language understanding data}
To maintain robust visual-language understanding, we sample 300,000 instances from the GQA dataset~\citep{hudson2019gqa}. We further augmented the dataset with 35,652 of GPT-4o–annotated description samples collected from our local environment.

\paragraph{Visual-language planning data}
We also include high-level planning data from RoboVQA~\citep{sermanet2023robovqa0}, EgoPlan~\citep{chen2023egoplan0bench0}, and ALFRED~\citep{shridhar2019alfred0}, which provide explicit step-by-step task decomposition and environment-aware planning supervision. The processed subset contains 241,603 data samples. 

\paragraph{Unsupervised movement prediction data}
To enhance spatial understanding and low-level motion grounding, we construct a small dataset of 10,575  samples in which we predict the movement transition between pairs of egocentric images. This unsupervised supervision allows the model learn spatial information without requiring manual annotation. 

\paragraph{DAgger experience data}
Finally, we incorporate on-policy trajectories collected through real-world executions with the DAgger algorithm~\citep{ross2010reduction}. We collected 70 successful traces with 3,629 \taskname{} training samples that span navigation in local environments, object-approach behaviors, and simple human-interaction tasks.

\subsection{Skill Setup}

For downstream skills, we first finetune a GROOT-N 1.5 model~\citep{gr00tn1_2025} to perform manipulation tasks, with details shown in Appendix~\ref{app:manipulation_training_details}. For locomotion, we adopt the official Unitree walking policy\footnote{
\url{https://github.com/unitreerobotics/unitree_sdk2}
} as the movement controller. We manually calibrate the robot's motion to achieve a positional precision of approximately 5 cm for forward/backward and lateral movements, and a turning precision of about 5 degrees. For speaking and querying behaviors, we currently assess task success by directly inspecting the model’s predicted natural-language outputs. The stand-up and crouch-down skills are implemented only in simulation, as the current Unitree locomotion policy does not support these actions in real-world deployment.

Actions for moving forward and turning left/right are merged, \ie they are not treated as discrete steps, to enhance motion speed and perform more human-like movement. 
To enable a faster movement of the robot, we amplify the forward distance predicted by the model by a factor of 1.2.

\section{Experiments}

\subsection{Experimental Setup}

\paragraph{Inference} For \projname{} inference, we use stochastic sampling with a temperature of 0.2. For all baseline models, we follow their original settings and apply greedy decoding. The instruction prompts we used for all different tasks are shown in Appendix~\ref{app:egoacting_prompts}.

\paragraph{Robot Setup} We deploy our model and all baseline methods on the same Unitree G1\footnote{\url{https://www.unitree.com/g1}} humanoid robot for real-world experiments. The robot is equipped with a pair of Unitree Dex3-1 hands\footnote{\url{https://www.unitree.com/Dex3-1}} and a custom 2-DoF head to support active perception\footnote{\url{https://github.com/BAAI-Agents/PAK/}}.
 A RealSense D455 camera\footnote{\url{https://www.realsenseai.com/products/real-sense-depth-camera-d455f/}}, as a typical camera model used in embodied projects, is
mounted on the custom head for \textbf{RGB-only} capture. We acquire 480p monocular RGB images from the camera and no depth data is leveraged in our model or experiments. 

\subsection{Baselines}
To investigate the hypothesis of leveraging existing navigation models for the movement-focused component of our proposed \taskname{}, we evaluate the navigation success rates of several representative navigation models based on vision–language foundation models on some of our benchmark tasks (including the movement part of Human Interaction, Traversability, and our virtual environment \taskname{} benchmark, all of which are introduced in Sections~\ref{sec:real-world-benchmark} and~\ref{sec:virtual-env-benchmark}). The baseline models considered are as follows:
\begin{enumerate}
    \item \textbf{NaVid~\citep{zhang2024navid0}} is a video-based large vision–language model for vision-and-language navigation that operates solely on monocular RGB video streams, achieving state-of-the-art map-free navigation and strong Sim2Real generalization.
    \item \textbf{Uni-NaVid~\citep{zhang2025uni0navid0}} is a video-based model that unifies multiple embodied navigation tasks within a single framework, enabling general-purpose, long-horizon navigation in unseen real-world environments.
    \item \textbf{NaVILA~\citep{cheng2024navila0}} is a two-level vision–language–action framework for legged robot navigation that converts language instructions into spatially grounded mid-level actions executed by a locomotion policy. We use its VLM component as a baseline.
\end{enumerate}



\begin{table}[t]
\caption{Single person human-robot interaction results comparing different models across three tasks.}
\label{tab:single_person_results_wide}
\centering
\resizebox{0.99\linewidth}{!}{
\begin{tabular}{l cccc}
\toprule
\multirow{2}{*}{\textbf{Model}} & \multicolumn{4}{c}{\textbf{Single Person Tasks}} \\
\cmidrule(lr){2-5}
 & \textbf{Approach} & \textbf{Say hi} & \textbf{Ask for location} & \textbf{Request items} \\
\midrule
NaVILA-7B   & 2/12  & - & - & - \\
NaVid-7B    & 8/12 & - & -& - \\
UniNaVid-7B & 8/12
 & - & -& - \\
 \midrule
EgoActor-4B & 12/12 & 12/12 & 12/12 & 11/12 \\
EgoActor-8B & \textbf{12/12} & \textbf{12/12} & \textbf{12/12} & \textbf{12/12} \\
\bottomrule
\end{tabular}
}
\end{table}


\begin{table}[t]
\caption{Multi-person human-robot interaction results for the ``Say Hi'' task using different model sizes.}
\label{tab:multiperson_egoactor_results}
\centering
\resizebox{0.99\linewidth}{!}{
\begin{tabular}{l ccccc}
\toprule
\multirow{2}{*}{\textbf{Model}} & \multicolumn{5}{c}{\textbf{Multi-person Attributes (Out-of-distribution)}} \\
\cmidrule(lr){2-6}
 & \textbf{Clothing} & \textbf{Accessories} & \textbf{Posture} & \textbf{Direction} & \textbf{Gender} \\
\midrule
EgoActor-4B & 8/12 & 7/12 & 8/12 & 11/12 & 10/12 \\
EgoActor-8B & \textbf{11/12} & \textbf{10/12} & \textbf{10/12} & \textbf{12/12} & \textbf{11/12} \\
\bottomrule
\end{tabular}
}
\end{table}


\begin{table}[t]

\caption{Unseen layout environment results on the \taskname{} Mobile Manipulation benchmark. Best results are bold.}
\label{tab:realworld_seen_manip_results}
\centering
\resizebox{0.49\textwidth}{!}{%
\begin{tabular}{llrrrrrrrrrrrrrrrrrrr}

\toprule
\multirow{5}{*}{\textbf{Seen Objects}}&\multirow{2}{*}{\textbf{Models}} & \multicolumn{2}{c}{\textbf{Approach and Pick}} & \multicolumn{2}{c}{\textbf{Approach and Place}}  \\
\cmidrule{3-4} \cmidrule{5-6} 
~ & ~ & Apple  & Bottle & Apple  & Bottle    \\
\cmidrule{2-6}
& \projname{}-4B & 5/6 & 5/6 & 3/6 & 4/6  \\
& \projname{}-8B & \textbf{5/6} & \textbf{6/6} & \textbf{6/6} & \textbf{6/6} \\
\midrule
\multirow{5}{*}{\textbf{Unseen Objects}} & \multirow{3}{*}{\textbf{Models}}  & \multicolumn{2}{c}{\textbf{Approach and Pick}} & \multicolumn{2}{c}{\textbf{Approach and Place}} \\

\cmidrule{3-4} \cmidrule{5-6} 
& & Pen Holder & Pink Cup & Pen Holder &  Pink Cup \\
\cmidrule{2-6}
& \projname{}-4B & 3/6 & 2/6 & 4/6 & 4/6 \\
& \projname{}-8B & \textbf{5/6} & \textbf{6/6} & \textbf{4/6} & \textbf{5/6} \\

\bottomrule
\end{tabular}}
\end{table}

\begin{table}[t]
\centering

\caption{Results on the proposed \taskname{} Traversability benchmark. Best results are bold.}
\label{tab:realworld_unseen_manip_results}
\resizebox{0.49\textwidth}{!}{%
\begin{tabular}{lrrrrrrrrrrrrr}

\toprule

\multirow{3}{*}{\textbf{Models}} & \multicolumn{4}{c}{\textbf{Seen Environments}} &  \multicolumn{4}{c}{\textbf{Unseen Environments}} \\
\cmidrule{2-5} \cmidrule{6-9} 
~ & \multicolumn{2}{c}{\textbf{Enter rooms}} & \multicolumn{2}{c}{\textbf{Leave rooms}} & \multicolumn{2}{c}{\textbf{Enter rooms}} & \multicolumn{2}{c}{\textbf{Leave rooms}}  \\
\cmidrule{2-3} \cmidrule{4-5}\cmidrule{6-7}\cmidrule{8-9}

~ & Left & Right & Left & Right & Left & Right & Left  & Right \\
\midrule
NaVILA-7B  & 5/12 & 4/12 & 3/12 & 3/12 & 2/8 & 1/8 & 3/8 & 1/8  \\
NaVid-7B & 3/12 & 5/12 & 9/12 & 8/12 & 1/8 & 0/8 & 8/8 & 4/8 \\
UniNaVid-7B & 4/12 & 3/12 & 6/12 & 4/12 & 2/8 & 0/8 & 5/8 & 5/8 \\
EgoActor-4B & 11/12 & 11/12 & \textbf{12/12} & 10/12 & 7/8 & 7/8 & 7/8 & 7/8 \\
EgoActor-8B & \textbf{11/12} & \textbf{12/12} & 10/12 & \textbf{10/12} & \textbf{7/8} & \textbf{7/8} & \textbf{8/8} & \textbf{7/8} \\

\bottomrule
\end{tabular}}

\end{table}


\begin{table*}[t]
\caption{Virtual environment results on our virtual benchmark. Best results are bold. ``m'' represents meters in the table.}
\label{tab:virtual_env_results}
\centering
\resizebox{0.95\textwidth}{!}{%
\begin{tabular}{lrrrrrrrrrr}

\toprule

\multirow{2}{*}{\textbf{Models}} & \multicolumn{8}{c}{\textbf{Distance to the Goal Position}} & \multirow{2}{*}{\textbf{\makecell[r]{Natural Language\\Action F1}}} & \multirow{2}{*}{\textbf{\makecell[r]{Final View\\Similarity}}} \\
\cmidrule{2-9}
~ & \textbf{$<$ 0.5 m} & \textbf{$<$ 0.8 m} & \textbf{$<$ 1.0 m} & \textbf{$<$ 1.2 m} & \textbf{$<$ 1.5 m} & \textbf{$<$ 2.0 m} & \textbf{$<$ 2.5 m} & \textbf{$<$ 3.0 m} \\
\midrule
NaVILA-7B  & 8.3\% & 21.0\% & 26.3\% & 28.8\% & 33.7\% & 41.5\% & 46.3\% & 52.2\% & - & 0.35 \\
NaVid-7B & 8.8\% & 15.1\% & 20.5\% & 23.9\% & 31.7\% & 42.0\% & 52.2\% & 60.0\% & - & 0.37 \\
UniNaVid-7B & 6.3\% & 15.6\% & 20.5\% & 23.9\% & 28.3\% & 35.1\% & 43.9\% & 51.7\% & - & 0.36\\
EgoActor-4B & 50.7\%&	63.7\%&	\textbf{70.6\%} &	74.1\% &	\textbf{78.9\%}	& \textbf{84.4\%}	& 86.5\%	& 87.8\% & 0.60 & \textbf{0.41} \\

EgoActor-8B & \textbf{51.4\%} &	\textbf{66.5\%}	& 69.9\% &	\textbf{74.1\%}	& 78.5\% & 	84.1\%	& \textbf{87.8\%}	& \textbf{89.9\%} & \textbf{0.62} & 0.41 \\

\bottomrule
\end{tabular}}

\end{table*}

\begin{figure}[t]
\centering
\includegraphics[width=0.95\linewidth]{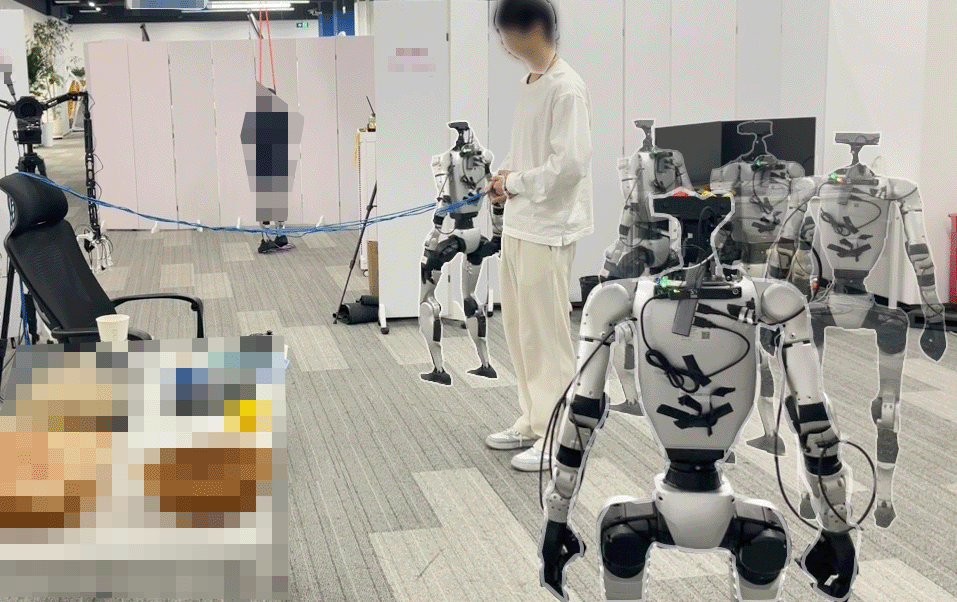} 
\caption{Multi-step illustration of obstacle avoidance generalization of our model, when faced with an unseen string obstacle.}
\label{fig:case_study_1}
\end{figure}

\begin{figure}[t]
\centering
\includegraphics[width=0.99\linewidth]{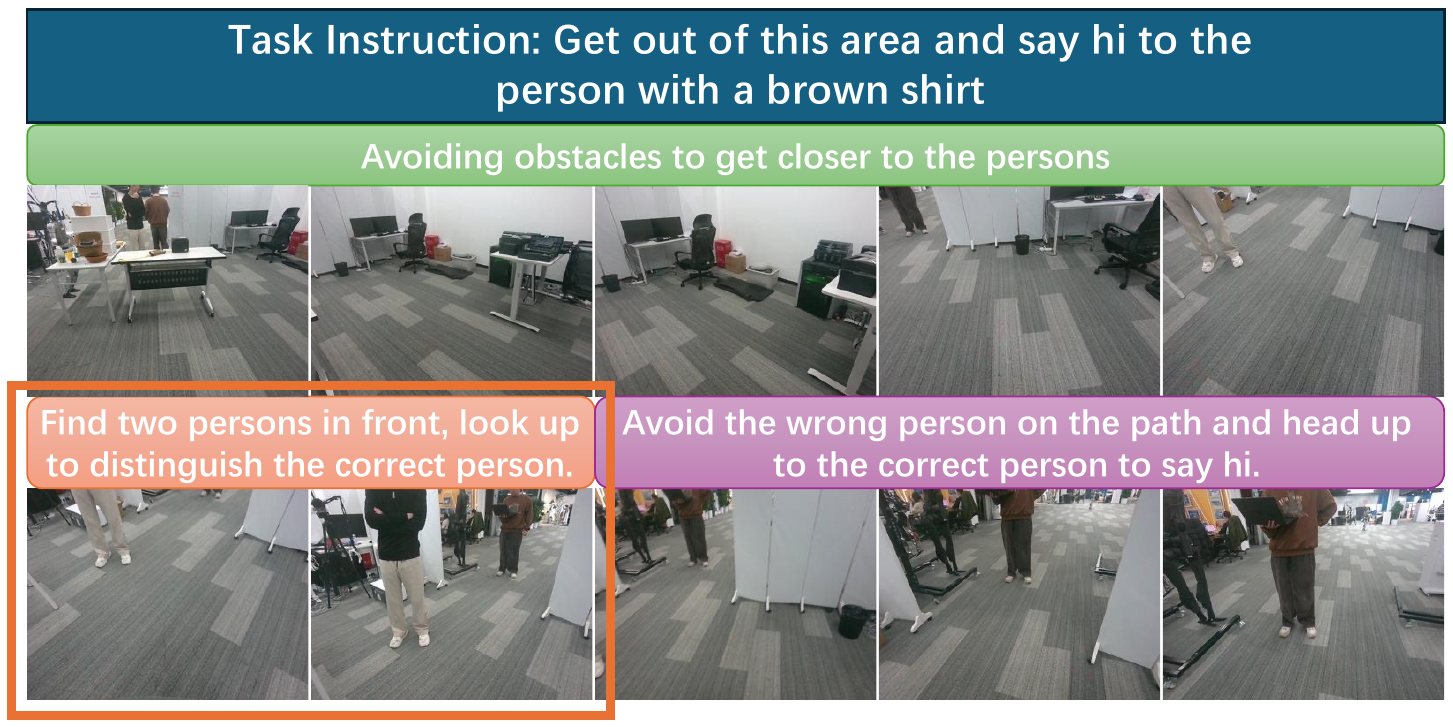} 
\caption{First-person view of an \projname{}'s active perception trace. Color description blocks highlight model's behaviors.}
\label{fig:case_study_2}
\end{figure}

\subsection{Real-world Benchmarking}
\label{sec:real-world-benchmark}


\paragraph{Human-robot Interaction}
In the human-robot interaction benchmark, the robot is required to navigate toward a specified person and perform the corresponding interaction, such as greeting, requesting information (\eg locations, etc.), or asking for help with an item.

\textbf{Setup.} All experiments are conducted in real-world environments and with people whose appearances and clothing are entirely different from those seen in the training data. For navigation-only baseline models, we evaluate whether the robot can stop at an acceptable location in front of a visible person (within approximately one meter and facing the person). For \projname{}, successful execution additionally requires generating an appropriate interaction action, such as requesting information.  
To further assess person-disambiguation capabilities, we design \textbf{\textit{unseen}} scenarios involving multiple individuals that differ in attributes such as clothing color, accessories, posture, facing direction, and gender. In each trial, the spatial arrangement of the individuals is randomized.

\textbf{Results.} As shown in Table~\ref{tab:single_person_results_wide} and \ref{tab:multiperson_egoactor_results}, both the 4B and 8B variants of \projname{} are generally able to guide the robot to approach a person and perform basic interactions. However, the 4B model shows weaker performance in scenarios with multiple people, particularly when fine-grained attribute-based identification is required. In contrast, the 8B model is able to identify the target person and carry out the instructed interaction in most tested cases. 
We emphasize that human interaction in our setting requires not only accurate navigation, but also the ability to translate task intent into appropriate body postures or dialogue, which we further examine through qualitative case studies in Appendix~\ref{app:human_interaction_case}.


\paragraph{Mobile Manipulation}
In the mobile manipulation benchmark, the robot is required to navigate to approach a target object and execute the corresponding manipulation action, such as picking or placing the object. An example of \projname{} controlling the robot to conduct a mobile manipulation task is shown in Appendix~\ref{app:mobile_manipulation}.

\textbf{Setup.} We evaluate the model in an unseen layout of the experimental environment, where desk, layout, and surrounding objects are arranged differently from those used during training. Target objects are placed at three distinct positions on the desk (left, center, and right), and each position is tested twice to account for stochasticity. 
For the \projname{} model, the evaluation includes both in-distribution and out-of-distribution object categories, whereas all object categories are in-distribution for the manipulation model.
Pick and place tasks are evaluated separately. 
Variations in object height are currently not considered in the real-world setting and are only analyzed in the virtual environment through additional case studies in Section~\ref{sec:case_study}.
In certain few cases, manipulation may fail due to lower-level execution policy issues, even when the target object is fully within the robot’s reachability; such cases are still counted as successful if manipulation was triggered at the correct moment and position. 

\textbf{Results.} We observe that the 8B \projname{} model is generally able to navigate to the correct objects and successfully support manipulation for both in-distribution and out-of-distribution objects under the unseen layout, indicating reasonable robustness to object and scene variations. For the 4B model, failures mainly occur when it predicts a manipulation action while still too far away from the target.

\paragraph{Traversability}
Traversability evaluates whether the robot can safely navigate through narrow spaces commonly encountered in daily environments without colliding with surrounding obstacles. As mentioned by \citet{internvla-n1}, most of the current VLM-based navigation models would suffer from hitting obstacles in the real-world environment. 
We evaluate these baseline models together with our \projname{}.

\textbf{Setup.} We focus on room entry and exit scenarios, as doorways are typically narrow and have been observed in preliminary tests to be particularly challenging and collision-prone for humanoid robots. The evaluation includes five real-world rooms, including three seen environments and two unseen ones. The evaluated rooms consist of three meeting rooms (seen during training), a private office, and a storage room, which differ in layout, visual style, and object arrangement. All room layouts are provided in Appendix~\ref{app:traversability}. For each room, we assess both entry and exit behaviors. To assess robustness to initial conditions, the robot is placed at 2 different starting positions for each trial—on the left and right sides of the doorway—and each position is tested 4 times.

\textbf{Results.} Quantitative results are reported in Table~\ref{tab:realworld_unseen_manip_results}. 
Our results show that \projname{} is generally able to traverse narrow passages and avoid collisions more reliably than baseline VLM-based navigation models. In contrast, existing VLN models often collide with door frames or nearby obstacles. We also observe that some baselines, such as NaVid, are generally effective at exiting rooms, but occasionally perform unnecessary rotations before door traversal, even when a straight path is available. These observations suggest that \projname{} demonstrates improved robustness in narrow-space movements. 
In addition, we conduct qualitative obstacle-avoidance experiments under unseen layouts and with unseen obstacles. These case studies further demonstrate the robustness of \projname{} in navigating narrow spaces and are discussed in detail in Section~\ref{sec:case_study} and Appendix~\ref{app:obstacle_avoidance}.

\subsection{Virtual Environment Benchmarking} 
\label{sec:virtual-env-benchmark}

\textbf{Setup.}
As described in Section~\ref{sec:data-acquisition}, we evaluate our \projname{} models using 205 labeled EgoActing samples collected in unseen virtual environments. Following the scene split protocol of the VLNCE dataset~\citep{krantz2020beyond}, all evaluation samples are drawn from environments unseen during training. Because LLM-based inference is stochastic, we evaluate each \projname{} model three times on the test set and report the average performance as the final result, with per-run results provided in Appendix~\ref{app:detail_3_results}. The evaluation results are summarized in Table~\ref{tab:virtual_env_results}.
For the evaluation metrics, we consider the precision of the ending position and the view similarity with the reference ending image. Additionally, we calculate the difference between the predicted natural language action and the reference natural language action by examining the F1 score of the 1-gram overlapping of them.

\textbf{Results.} As shown in Table~\ref{tab:virtual_env_results}, EgoActor generalizes well to unseen environments and target objects, with the 8B model performing slightly better under smaller distance thresholds, while the 4B and 8B models achieve overall comparable performance. We check the failure cases and find that most errors arise from ambiguous labelled instructions or visually degraded or blurry virtual environments, with additional failures occurring in unfamiliar scene types such as churches or historical sites that differ substantially from the training data. 

For baseline models, we observe that under the standard VLN success criterion (\textbf{$<$3.0 m}), performance remains comparable to their reported VLNCE Room-to-Room results (around 50\%). However, under stricter criteria requiring precise positioning for interaction, these models frequently fail to stop at appropriate locations, often hallucinating continued navigation instead of stopping to execute the intended interaction. 

\begin{figure}[t]
\centering
\includegraphics[width=0.99\linewidth]{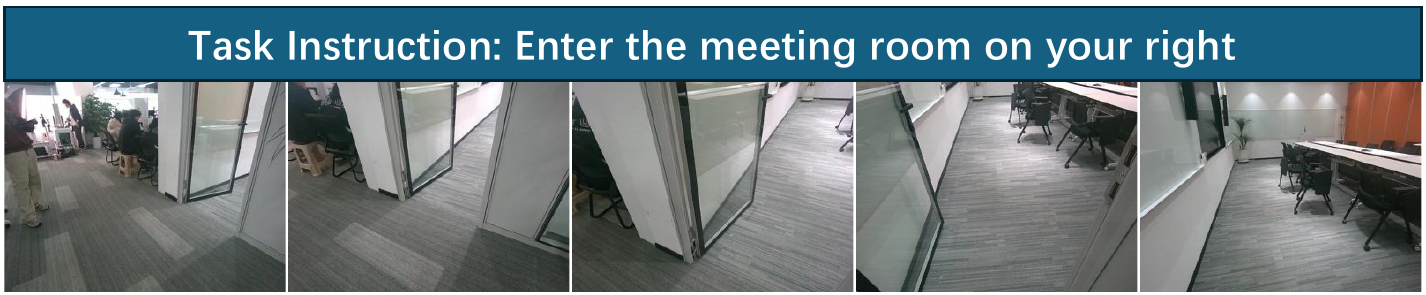} 
\caption{First-person view of an \projname{}'s traversability trace, showing the robot walking through a doorway.}
\label{fig:case_study_3}
\end{figure}

\begin{figure}[t]
\centering
\includegraphics[width=0.99\linewidth]{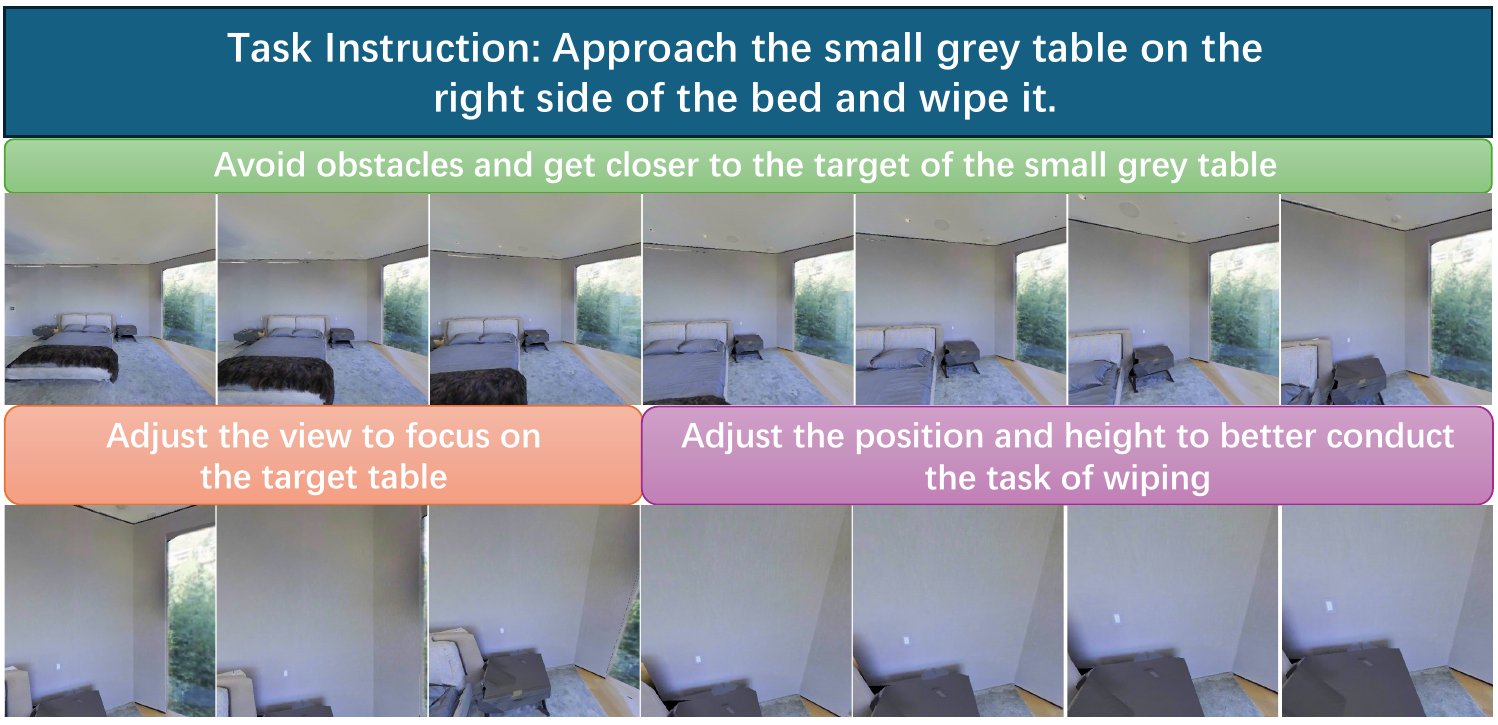} 
\caption{First-person view of an \projname{}'s height change ability trace in virtual environments. Color description blocks highlight model's behaviors.}
\label{fig:case_study_4}
\end{figure}

\subsection{Case Studies}
\label{sec:case_study}
In this section, we discuss representative case studies that demonstrate \projname{}’s capabilities in obstacle avoidance, active perception, spatial understanding, and human-like movement. A detailed example of human interaction is provided in Appendix~\ref{app:human_interaction_case}.
\paragraph{Obstacle Avoidance} 
Through all real-world experiments, the G1 robot rarely collides with obstacles. In a couple instances, minor side collisions occur when the robot focuses on avoiding a large obstacle ahead and temporarily loses sight of small obstacles that have previously passed its view. 
Moreover, as illustrated in Fig.~\ref{fig:case_study_1}, the model can generalize and handle unseen obstructions, like the rope/stripe, and find a way to walk around them.

\paragraph{Active Perception}
We observe that the robot looks downward to verify obstacle positions while passing them, improving obstacle-avoidance success. Toward the end of a trajectory, the model also actively keeps its gaze fixed on the target object to enable a smoother transition to manipulation or interaction actions. The case shown in Fig.~\ref{fig:case_study_2} illustrates that the model actively moves backwards and looks upward to identify the color of upper-body clothing when only the lower bodies of two people are initially visible, as the instruction requires greeting the person wearing a specific shirt.

\paragraph{Spatial Understanding}
Unlike VLMs trained primarily in simulation, our model learns from human videos and thus develops stronger spatial understanding. For example, it predicts different forward-movement distances when encountering a clear, wide path versus a partially obstructed path, or negotiating a doorway corner. Its predicted turning angles also adapt appropriately across different spatial configurations. An example of this in a mobile manipulation scenario is shown in Appendix~\ref{app:mobile_manipulation}.

\paragraph{Human-like Behaviors} 
Trained on real-world videos, the model naturally exhibits human-like movement behaviors. It may move backward when too close to obstacles or after completing a manipulation to reorient toward the next target. During corner turning, the model often combines forward action, turning, and strafing (\eg turning left while moving forward and strafing right), mirroring how humans execute smooth turns while maintaining visual alignment with the path (see Fig.~\ref{fig:case_study_2} and Fig.~\ref{fig:case_study_3}). Height adjustments are also observed in the virtual environment experiments (see Fig.~\ref{fig:case_study_4}).



\section{Conclusion}
We propose \projname{}, a unified vision–language model that addresses overlooked challenges in grounding high-level task intentions into egocentric, executable multi-step actions for humanoid robots in real-world settings, without requiring extra sensing modalities, multiple cameras, nor extensive teleoperation. By jointly predicting locomotion, manipulation, human interaction, and head movements, \projname{} tightly integrates perception and execution in dynamic environments. Trained on easily scalable diverse real-world, spatial reasoning, and simulated data, the model demonstrates strong generalization and timely inference. Extensive evaluations in both simulation and physical robots show \projname{} effectively bridges abstract task planning and low-level action execution, offering a practical step toward scalable humanoid autonomy. \projname{} is to be released as an open testbed to support future research (including code, models, dataset, and benchmark). Current limitations are further discussed in Appendix~\ref{app:limitions}.
\bibliographystyle{plainnat}
\bibliography{references}

\clearpage
\tableofcontents  

\section{Limitations}
\label{app:limitions}
The effectiveness of the proposed \projname{} relies heavily on the reliability of external components, including high-level planners (\eg large language models) and downstream skills such as locomotion policies and visual-language-action models for manipulation. On its own, \projname{} does not function as a fully end-to-end system, as it depends on these supporting models. Future work could explore integrating these capabilities into a single, unified framework to facilitate more seamless deployment on humanoid robots.  Another limitation lies in its handling of long-term context: the model may occasionally fall into locally optimal but incorrect decision patterns when navigating extended or multi-stage tasks.

\section{Data Format}
\label{app:data_format}
To enable robust grounding from high-level instructions to executable humanoid actions, we design a structured EgoActing prompt that explicitly exposes the model to egocentric visual context, temporal history, and recent action–observation pairs. The prompt frames the model as a vision–language agent operating from a first-person perspective and requires it to reason over both long-term historical observations and short-term recent frames. By constraining the output to a predefined set of low-level locomotion, perception, manipulation, and interaction primitives, the prompt encourages spatially grounded, temporally coherent decision-making while preventing unconstrained or hallucinated responses. This design allows the model to infer the next usable action step conditioned on instruction intent, environmental state, and execution history, closely aligning the inference process with real-world humanoid control requirements.

\begin{tcolorbox}[
    colback=purple!5!white, 
    colframe=purple!75!black, 
    title=EgoActing Prompt Design, 
    fonttitle=\bfseries, 
    boxrule=0.5mm, 
    arc=2mm, 
    left=2mm, 
    right=2mm, 
    top=2mm, 
    bottom=2mm, 
    breakable
]

You are a Vision Language Model specialized in processing the first person view images of embodied robots.\\
Your task is to analyze the provided image and respond to queries with answers. Focus on the spatial relations in the image and make the right decisions. \\

Given the following instruction, a series of sampled historical observation and recent observation image frames, predict a usable action sequence that you should perform next. Output format: 'Turn [direction] [degrees] degrees; Look [direction] [degrees] degrees; Move [direction] [distance] meters; [direction] sidewalk [distance] meters; [manipulation action text]; [interaction action text]; Stop and no action'. 

Your task is: 
\begin{verbatim}
[instruction] 
\end{verbatim}
\vspace{1em}

Sampled Historical Observations:

\begin{verbatim}
[Sampled Historical Observation #1] 
[Sampled Historical Observation #2] 
[Sampled Historical Observation #3] 
[Sampled Historical Observation #4] 
[Sampled Historical Observation #5] 
[Sampled Historical Observation #6] 
[Sampled Historical Observation #7] 
[Sampled Historical Observation #8] 
[Sampled Historical Observation #9] 
[Sampled Historical Observation #10] 
\end{verbatim}
\vspace{1em}

Recent Observations:\\

\begin{verbatim}
[Recent Observation #1] 
\end{verbatim}
Next action:
\begin{verbatim}
[Recent Action #1] 
\end{verbatim}
\vspace{1em}

\begin{verbatim}
[Recent Observation #2] 
\end{verbatim}
Next action:
\begin{verbatim}
[Recent Action #2] 
\end{verbatim}
\vspace{1em}

\begin{verbatim}
[Recent Observation #3] 
\end{verbatim}
Next action:





    
    
\end{tcolorbox}

\section{Data Processing Details}
\label{app:data_processing}

\subsection{EgoTaskQA Data Processing}
\label{app:egotaskqa}

EgoTaskQA is a benchmark that evaluates models' event understanding capabilities with goal-oriented questions. The dataset provides fine-grained temporal annotations, including the start and end frames of atomic actions, which makes it suitable for constructing sequential decision-making data. To adapt EgoTaskQA into the \textit{\taskname{}} format for training vision--language models, we reorganize each annotated sequence into structured action-conditioned observation samples.

\paragraph{Overview}
Each processed training sample consists of three components: 
\begin{enumerate}
    \item a global natural language instruction 
    \item a set of historical observations 
    \item three interleaved recent observation-action pairs
\end{enumerate}

This structure is designed to accommodate  \taskname{}, where models must predict the next feasible action based on the historical and recent observations.

\paragraph{Instruction Construction}
EgoTaskQA does not provide explicit instruction annotations; instead, it includes detailed annotations of distinct action steps within each video. We first concatenate three adjacent actions (or fewer if insufficient actions are available) into a single instruction using a predefined set of connective patterns (\eg “and then,” “and next,” “continue to”). For example, an instruction such as “Get the tank from the table, and then open the tank” contains two sequential manipulation actions. We define the start frame as 60 frames before the start frame of the first action, unless there are preceding actions or this offset exceeds the beginning of the video.


\paragraph{Recent Observation-Action Pairs}
We start with a target frame that is supposedly used to train the model to predict the next action.
For each target frame, we construct three temporally adjacent observation--action pairs by traversing the sequence backward with a fixed stride of five frames. 
The corresponding action is determined according to the frame's temporal role:
\begin{enumerate}
    \item if the frame corresponds to the start of a manipulation phase, the action is the extracted manipulation action;
    \item if the frame lies within a movement segment, the action is constructed by aggregating the calculated camera pose difference between frames;
    \item if the frame follows the completion of the final manipulation step, a special \textit{Stop and no action} token is used.
\end{enumerate}


\paragraph{Navigation Action Aggregation and Filtering}
Navigation actions are derived by accumulating camera pose changes (\eg rotation, translation, vertical motion) between consecutive frames. The changes along opposite directions are algebraically combined to account for cancellation effects. To ensure action saliency, we apply magnitude-based thresholds (5 degrees for angular motion and 0.1 meters for translational motion) to determine whether a navigation action is considered valid. 


\paragraph{Historical Observations}
Historical observations provide long-term visual context. For each sample, we collect all frames between the instruction start frame and the first recent observation frame, regardless of phase boundaries. From this interval, ten frames are uniformly sampled to form the historical observation set. This strategy avoids reliance on phase-specific image lists and ensures consistent temporal coverage.

\paragraph{Sample Variants}
For manipulation actions, we construct two types of samples: one where the target action is a manipulation instruction, and another where the target action is the \textit{Stop and no action} token, indicating task completion.

\paragraph{Final Dataset}
Following the above procedure, EgoTaskQA sequences are converted into structured \taskname{} samples with aligned visual context, recent action history, and explicit action supervision.
This processed dataset enables training models to jointly reason over instruction context, recent egocentric observations, and temporally grounded actions. The raw dataset contains 6,599,590 samples, of which 531,090 include natural language action predictions. Due to computational constraints, we randomly sample 40,000 instances from the full set and 120,000 instances from the subset containing natural language action predictions, resulting in 160,000 training samples for \taskname{}.

\subsection{Virtual Environment Data Processing}
\label{app:virtual_env_processing}

This section describes the processing pipeline used to convert labeled virtual-environment trajectories into training and evaluation samples compatible with \taskname{}-style action prediction.

\paragraph{Trajectory Parsing}
Each episode is stored as a trajectory consisting of RGB observations and discrete low-level actions. We first normalize action names (\eg \texttt{MOVE\_FORWARD}, \texttt{TURN\_LEFT}, \texttt{LOOK\_UP}) and append a terminal \texttt{STOP} action to mark episode completion. For each step, the corresponding RGB image path is recorded, forming an aligned image–action sequence.

\paragraph{Action Merging}
To better reflect continuous robot motion and reduce action sparsity, we merge pairs of consecutive low-level actions into a single executable instruction. The merging process aggregates rotations, forward motion, lateral strafing, and camera pitch adjustments, while introducing small randomized perturbations to movement distances and angles to improve robustness. Terminal actions are mapped to a unified \texttt{Stop and no action} command.

\paragraph{Sliding-Window Sample Construction}
From each episode, we construct multiple training samples using a sliding-window strategy. For a given window position, three recent observation–action pairs are selected at fixed temporal intervals, along with all preceding images treated as historical context. This design enables the model to jointly reason over long-term visual history and short-horizon action prediction.

\paragraph{Final Sample Format}
Each sample consists of (1) a natural-language instruction, (2) a sequence of recent image–action pairs, and (3) a set of historical observation images. All samples are serialized into a unified JSON format for downstream training and evaluation.

This pipeline yields temporally coherent, instruction-conditioned samples that closely match the inference-time setting of EgoActor in virtual environments.

\section{Supported Skills}
\label{app:supported_skills}
Table~\ref{tab:skills} summarizes the set of skills supported by \projname{}, covering movement, perception, manipulation, and human–robot interaction. We represent these skills using two complementary forms of language-based actions: \emph{structured language actions} for spatial control and \emph{natural language actions} for open-ended interaction.

\paragraph{Structured language actions}
Movement and active perception are modeled using structured, interpretable action templates that explicitly specify the action type, direction, and magnitude (\eg \textit{Turn left 30.0 degrees}, \textit{Move forward 0.26 meters}). These actions support egocentric locomotion, body-height adjustment, lateral motion, and head orientation, enabling precise spatial positioning from visual observations. Small-magnitude motions are filtered to reduce noise and instability.

\paragraph{Natural language actions}
Manipulation and human-interaction skills are represented using natural language actions rather than a fixed action inventory. This includes object manipulation (\eg picking, placing, opening), communicative behaviors (\eg speaking, asking), and gesture-based interactions. This representation allows flexible composition, generalization to unseen actions, and direct grounding of task intent into executable language commands.

We suggest that the natural language actions could be easily extended to facilitate more varied actions in the future. 

\begin{table*}[t]
\caption{Supported skills in our training datasets.}
\label{tab:skills}
\centering
\resizebox{0.8\textwidth}{!}{%
\begin{tabular}{lll}
\toprule
\textbf{Skill Category} & \textbf{Skill Description} & \textbf{Action Example} \\
\midrule
\multicolumn{3}{c}{\bf Structured Language Skills} \\
\midrule
\multirow{5}{*}{Movement Skills} 
 & Move forward/backward & \texttt{Move forward/backward 0.26 meters} \\
 & Turn left/right & \texttt{Turn left/right 30.0 degrees} \\
 & Strafe left/right & \texttt{Left/right sidewalk 0.40 meters} \\
 & Stand up & \texttt{Rise up 0.12 meters} \\
 & Crouch down & \texttt{Lower down 0.08 meters} \\
\midrule
 Active Perception Skills & Look up/down & \texttt{Look up/down 10.0 degrees} \\
 \midrule
 \multicolumn{3}{c}{\bf Natural Language Skills}  \\
\midrule
\multirow{4}{*}{Human Interaction Skills} & Confirm/denial gesture & \texttt{Confirm with the woman in front of you} \\
 & Say hi & \texttt{Say hi to the boy} \\
 & Speak & \texttt{Speak ``How you doing?''} \\
 & Ask & \texttt{Ask ``Where is the bathroom?''} \\
 \midrule
\multirow{11}{*}{Manipulation Skills} & Grab/Grasp/Pick up | & \texttt{Pick up the water bottle} \\
 & Pull |& \texttt{Pull the drawer}  \\
 & Place | on | & \texttt{Place the plate on the desk} \\
 & Open |& \texttt{Open the door}  \\
 & Close |& \texttt{Close the door}  \\
 & Wash & \texttt{Wash hands} \\
 & Pour from | into  | & \texttt{Pour from the bottle into the cup} \\
 & Turn on |  & \texttt{Turn on the washing machine} \\
 & Turn off | & \texttt{Turn off the lamp} \\
 & Point to | & \texttt{Point to the painting} \\
 & Drop | & \texttt{Drop the garbage} \\
 \bottomrule
\end{tabular}}

\end{table*}

\section{Training Details of the Manipulation Model}
\label{app:manipulation_training_details}
We conducted full fine-tuning of the GROOT-N 1.5 model~\citep{gr00tn1_2025} for 40,000 training steps using an 80GB A800 GPU with a batch size of 50 with the official implementation\footnote{\url{https://github.com/NVIDIA/Isaac-GR00T}}. For each task, we expanded the task descriptions to multiple different versions and randomly sampled one description at each training iteration to increase linguistic diversity. Data were collected using a monocular RGB camera. During acquisition, all objects were placed on a white table with a height of 70 cm. The grasped objects included apples, water bottles, plastic cups, oranges, tissue boxes, pen holders, and bowls. Containers consisted of both square and round plates, as well as shallow baskets. In total, the dataset comprises approximately 700 samples.

\section{Detailed Results for All the Three Runs }
\label{app:detail_3_results}

We report the per-run evaluation results corresponding to the averaged performance presented in the main paper. To account for stochasticity introduced by LLM-based inference, we conduct three independent evaluation runs under identical settings. Tables~\ref{tab:three_runs_4b} and~\ref{tab:three_runs_8b} summarize the success rates at different distance thresholds for each run and different models, enabling a finer-grained comparison of performance stability and variance across runs and model scales.

\begin{table*}[t]
\centering
\caption{Multi-threshold success rates across three evaluation runs for the 4B \projname{} model.}
\label{tab:three_runs_4b}
\resizebox{1.0\textwidth}{!}{%
\begin{tabular}{ccc}

\begin{tabular}{lcc}
\toprule
\multicolumn{3}{c}{\textbf{Run 1}} \\
\midrule
\textbf{Distance} & \textbf{Success} & \textbf{Count} \\
\midrule
0.5 m  & 52.20\% & 107/205 \\
0.8 m  & 62.93\% & 129/205 \\
1.0 m  & 70.24\% & 144/205 \\
1.2 m  & 73.17\% & 150/205 \\
1.5 m  & 78.05\% & 160/205 \\
2.0 m  & 84.39\% & 173/205 \\
2.5 m  & 86.34\% & 177/205 \\
3.0 m  & 87.32\% & 179/205 \\
\bottomrule
\end{tabular}
&
\begin{tabular}{lcc}
\toprule
\multicolumn{3}{c}{\textbf{Run 2}} \\
\midrule
\textbf{Distance} & \textbf{Success} & \textbf{Count} \\
\midrule
0.5 m  & 48.29\% & 99/205 \\
0.8 m  & 62.93\% & 129/205 \\
1.0 m  & 68.78\% & 141/205 \\
1.2 m  & 74.15\% & 152/205 \\
1.5 m  & 80.00\% & 164/205 \\
2.0 m  & 84.39\% & 173/205 \\
2.5 m  & 87.80\% & 180/205 \\
3.0 m  & 88.78\% & 182/205 \\
\bottomrule
\end{tabular}
&
\begin{tabular}{lcc}
\toprule
\multicolumn{3}{c}{\textbf{Run 3}} \\
\midrule
\textbf{Distance} & \textbf{Success} & \textbf{Count} \\
\midrule
0.5 m  & 51.71\% & 106/205 \\
0.8 m  & 65.37\% & 134/205 \\
1.0 m  & 72.68\% & 149/205 \\
1.2 m  & 75.12\% & 154/205 \\
1.5 m  & 78.54\% & 161/205 \\
2.0 m  & 84.39\% & 173/205 \\
2.5 m  & 85.37\% & 175/205 \\
3.0 m  & 87.32\% & 179/205 \\
\bottomrule
\end{tabular}

\end{tabular}
}
\end{table*}

\begin{table*}[t]
\centering
\caption{Multi-threshold success rates across three evaluation runs for the 8B EgoActor model.}
\label{tab:three_runs_8b}
\resizebox{1.0\textwidth}{!}{%
\begin{tabular}{ccc}

\begin{tabular}{lcc}
\toprule
\multicolumn{3}{c}{\textbf{Run 1}} \\
\midrule
\textbf{Distance} & \textbf{Success} & \textbf{Count} \\
\midrule
0.5 m  & 48.78\% & 100/205 \\
0.8 m  & 66.34\% & 136/205 \\
1.0 m  & 68.78\% & 141/205 \\
1.2 m  & 73.17\% & 150/205 \\
1.5 m  & 78.05\% & 160/205 \\
2.0 m  & 83.41\% & 171/205 \\
2.5 m  & 86.83\% & 178/205 \\
3.0 m  & 88.78\% & 182/205 \\
\bottomrule
\end{tabular}
&
\begin{tabular}{lcc}
\toprule
\multicolumn{3}{c}{\textbf{Run 2}} \\
\midrule
\textbf{Distance} & \textbf{Success} & \textbf{Count} \\
\midrule
0.5 m  & 51.71\% & 106/205 \\
0.8 m  & 66.34\% & 136/205 \\
1.0 m  & 68.78\% & 141/205 \\
1.2 m  & 73.66\% & 151/205 \\
1.5 m  & 77.07\% & 158/205 \\
2.0 m  & 84.39\% & 173/205 \\
2.5 m  & 88.29\% & 181/205 \\
3.0 m  & 90.73\% & 186/205 \\
\bottomrule
\end{tabular}
&
\begin{tabular}{lcc}
\toprule
\multicolumn{3}{c}{\textbf{Run 3}} \\
\midrule
\textbf{Distance} & \textbf{Success} & \textbf{Count} \\
\midrule
0.5 m  & 53.66\% & 110/205 \\
0.8 m  & 66.83\% & 137/205 \\
1.0 m  & 72.20\% & 148/205 \\
1.2 m  & 75.61\% & 155/205 \\
1.5 m  & 80.49\% & 165/205 \\
2.0 m  & 84.39\% & 173/205 \\
2.5 m  & 88.29\% & 181/205 \\
3.0 m  & 90.24\% & 185/205 \\
\bottomrule
\end{tabular}

\end{tabular}
}
\end{table*}

\section{Traversability Scenes}
\label{app:traversability}
\begin{figure*}[t]
\centering
\includegraphics[width=0.95\linewidth]{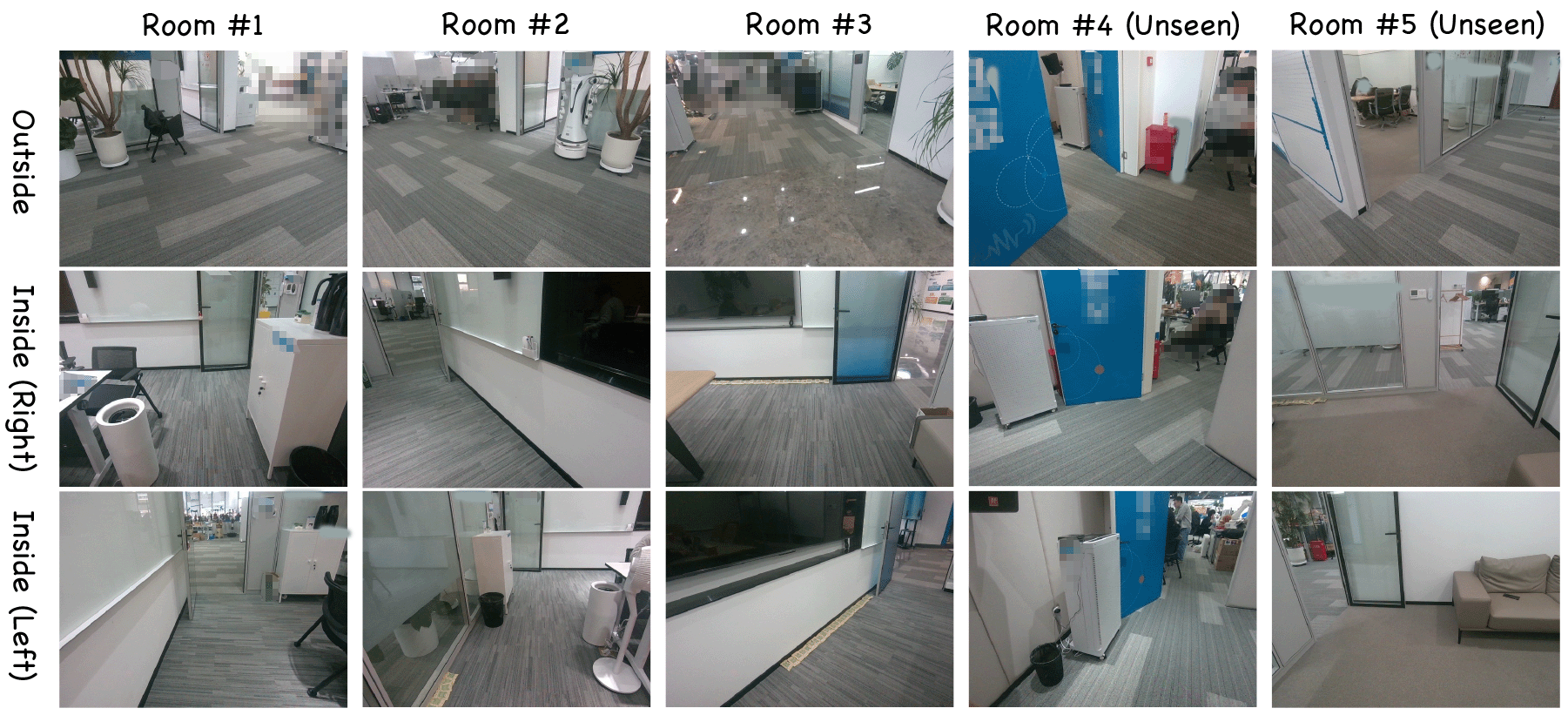} 
\caption{An illustration of the different scenes we used in our traversability experiments.}
\label{fig:traversability_scenes}
\end{figure*}

Fig.~\ref{fig:traversability_scenes} illustrates the real-world environments used in the traversability evaluation. Traversability focuses on assessing whether a humanoid robot can safely navigate through narrow spaces—particularly doorways—without colliding with surrounding obstacles, which is a common failure mode for vision–language–model-based navigation systems in real-world settings~\citep{internvla-n1}.

We evaluate traversability using room entry and exit scenarios, as doorways are typically constrained in width and require precise control and obstacle awareness. The evaluation includes five real-world rooms: three meeting rooms that are seen during training, and two unseen environments consisting of a private office and a storage room. These rooms differ in layout, visual appearance, carpet, and object arrangement, providing diverse traversal conditions.

For each room, both entering and exiting behaviors are tested. To assess robustness to initial positioning, the robot starts from two different locations relative to the doorway (left and right), with four repeated trials per starting position. This setup allows systematic evaluation of collision avoidance and narrow-passage traversal across varying spatial configurations.

\section{Additional Case Study}
We provide additional videos in the supplemental materials to illustrate \projname{}’s behavior across diverse scenarios. Occasional latency in the videos is primarily due to network instability rather than the model’s inference speed. The model itself operates with sub-second action prediction, while the observed end-to-end delay is mostly affected by transmission and streaming conditions.
\subsection{Mobile Manipulation Case Study}
\label{app:mobile_manipulation}
\begin{figure*}[t]
\centering
\includegraphics[width=0.8\linewidth]{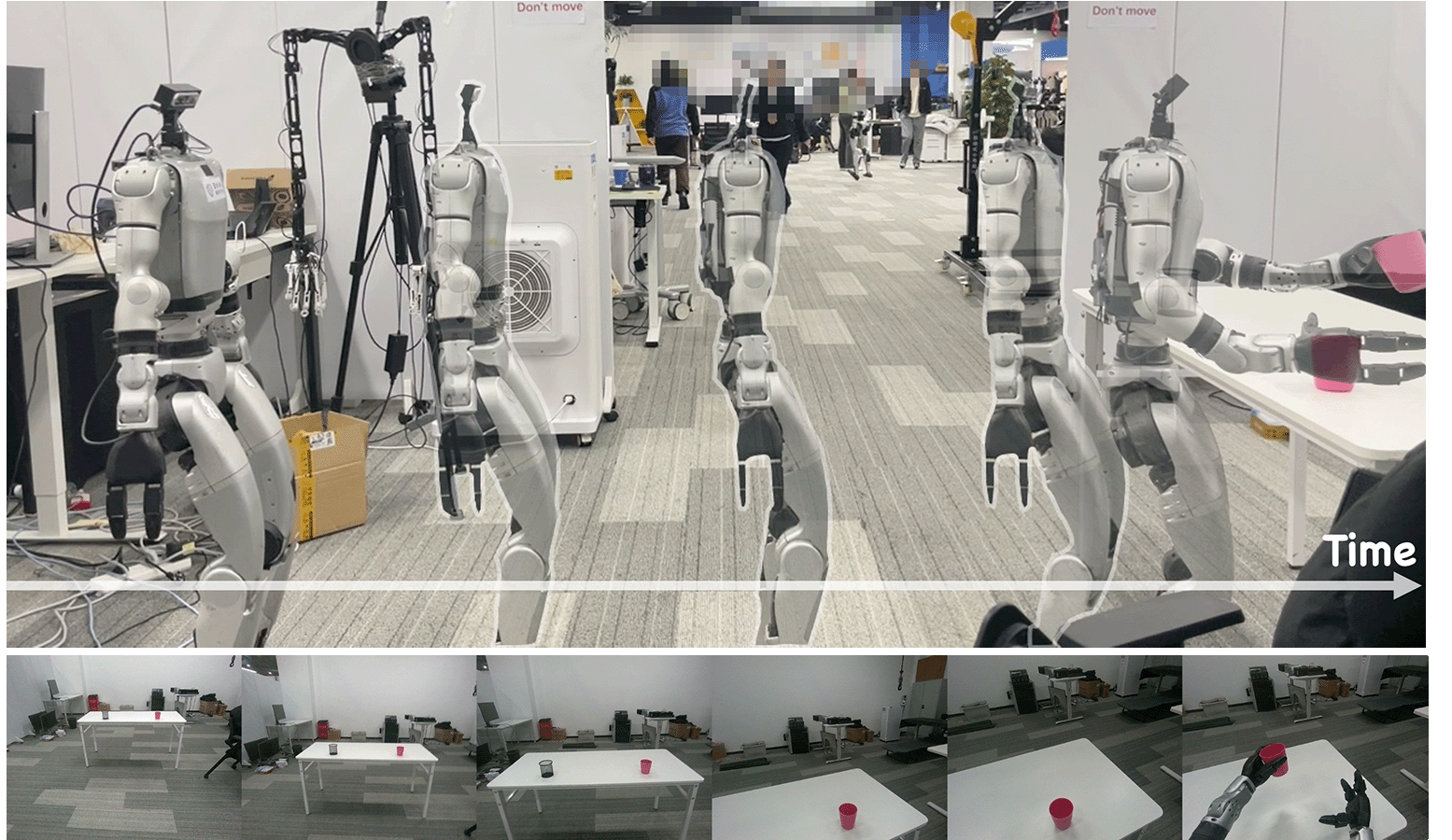} 
\caption{An illustration of our model conducting the mobile manipulation task: ``Approach and grab the pink cup''.}
\label{fig:case_study_manipualtion}
\end{figure*}

Fig.~\ref{fig:case_study_manipualtion} illustrates \projname{} performing a mobile manipulation task. We also provide videos of the illustrations in the supplemental materials. Starting from a distant position, the robot first takes larger locomotion steps to efficiently approach the workspace and progressively adjusts its trajectory to align with the target. As it nears the object, the robot transitions to smaller, fine-grained movements to precisely refine its position for manipulation. The example further demonstrates \projname{}’s ability to adapt its motion to the spatial configuration of the scene: the target object—a previously unseen pink cup—is picked up successfully despite the presence of another object (a pen holder) on the desk, highlighting robust spatial reasoning and fine positioning under cluttered conditions.

\subsection{Obstacle Avoidance}
\label{app:obstacle_avoidance}
We provide additional qualitative examples of obstacle avoidance behaviors in the supplemental videos. These cases illustrate the model’s ability to navigate around static and dynamic obstacles under unseen layouts, particularly in narrow spaces. The videos highlight how \projname{} adjusts its locomotion primitives to maintain safe clearance while preserving task progress.

\subsection{Human-robot Interaction} 
\label{app:human_interaction_case}
Table~\ref{tab:case_nla} presents representative examples of natural language actions predicted by \projname{} in human-interaction scenarios. Given high-level instructions with varying intents, the model generates context-appropriate verbal actions, such as asking for directions, requesting information, or initiating polite inquiries. Note that this task is trained with fewer than 20 samples in the training data. These examples suggest that \projname{} can flexibly map diverse instructional intents to corresponding linguistic actions, enabling basic information-seeking and communicative behaviors during human–robot interaction.

\begin{table*}[t]
\caption{An example of the predicted natural language actions that show the \projname{} could transform the intentions in the instructions into actual words.}
\label{tab:case_nla}
\centering
\resizebox{0.99\textwidth}{!}{%
\begin{tabular}{l|l}
\toprule
\textbf{Instructions} & \textbf{Predicted Natural Language Actions }\\
\midrule
 Approach the person and ask him the way to cook Kung Pao Chicken & Ask ``How do you cook Kung Pao Chicken?'' \\
 Approach the person and ask him to show you the way to the reception & Ask ``Could you please show me the way to the reception?'' \\
 Approach the person and ask the status of the air conditioner & Ask ``How is the working of the air conditioner?'' \\
Approach the person and ask him to check the news politely & Ask ``Could you please check the news?'' \\
Approach and ask the person to check the news & Ask ``Do you know the news?'' \\
Approach and ask the person to check the news & Ask ``check the news'' \\
Approach the person and ask him the status of the data labeling & Ask ``How is the progress of the data labeling?'' \\
Approach the person and ask him to hand you the flowers & Ask ``Could you please give me the flowers?'' \\

\bottomrule
\end{tabular}}

\end{table*}

\section{Data Samples for the EgoActing Task}
\label{app:egoacting_sample}
Table~\ref{tab:instruction_action_examples} illustrates representative instruction–action pairs from both real-world and virtual environments for the EgoActing task. Each instruction specifies a high-level goal, while the corresponding natural language action describes the concrete behavior executed by the agent. The examples highlight the diversity of tasks, including navigation, object interaction, and human interaction, and demonstrate how the dataset captures both real-world complexities and controlled virtual scenarios. This paired structure supports training and evaluation of models capable of grounding language instructions into egocentric, low-level action sequences.

\begin{table*}[t]
\centering
\caption{Examples of instruction–action pairs in real-world and virtual environments.}
\label{tab:instruction_action_examples}
\resizebox{\textwidth}{!}{
\begin{tabular}{ll}
\toprule
 \textbf{Instructions} & \textbf{Natural Language Actions} \\
\midrule
\multicolumn{2}{c}{\textbf{Real-World Environments}}\\
\midrule
 Turn a large right to get out of this area and say hi to the robot in a blue shirt.
& Say hi to the robot \\
 Get out of this area and turn a large left to the leftmost hallway.
& Stop and no action \\
Turn left and go straight to a kitchen, stop in front of the people.
& Stop and no action \\
Search and approach the girl and ask her information about humanoid robots.
& Ask ``do you know anything about humanoid robots'' \\
Approach and grab the toy bear on the white table.
& Grab the toy bear \\
\midrule
\multicolumn{2}{c}{\textbf{Virtual Environments}} \\
\midrule
Turn right until you see a window, then point to it.
& Point to the window \\
Walk forward to the chair and pull out the red chair in the middle.
& Pull out the red chair in the middle \\
Approach and put the towels on the bathtub.
& Put the towels \\
Approach the lamp on the left side of the bed and turn on the lamp.
& Turn on the lamp \\
Approach and clean the mirror on the wooden cabinet.
& Clean the mirror \\
\bottomrule
\end{tabular}
}
\end{table*}

\section{Prompts for Different Tasks}
\label{app:egoacting_prompts}

To evaluate \projname{} across a diverse set of embodied capabilities, we design task-specific instruction prompts that are high-level yet explicit, ensuring that the intended goal and required actions are clearly specified without prescribing low-level motor details. Specifically,

Table~\ref{tab:human_interaction_prompts} presents examples of human-interaction prompts, including greeting, asking for information, and requesting objects. These prompts cover both single-person and multi-person settings and require the model to resolve referential ambiguity using visual attributes (\eg clothing, gestures, relative position) or relational descriptions.

Table~\ref{tab:task_prompts} summarizes the instruction prompts used for mobile manipulation tasks. These prompts focus on pick-and-place behaviors with varying objects and appearances, requiring the robot to approach the workspace, localize the target, and execute manipulation actions under egocentric observations.

Finally, Table~\ref{tab:room_prompts} lists prompts for room-level tasks, such as entering or exiting rooms and performing simple interactions after navigation. These prompts are designed to assess the model’s ability to handle narrow passages, spatial transitions, and action sequencing in real-world indoor environments.

\begin{table*}[t]
\caption{Examples of human-interaction task prompts used in our evaluation.}
\label{tab:human_interaction_prompts}
\centering
\resizebox{\textwidth}{!}{%
\begin{tabular}{lll}
\toprule
\textbf{Task Category} & \textbf{Setting} & \textbf{Prompt} \\
\midrule
\multirow{15}{*}{Say Hi}
& Single Person & Approach the person with the grey sweater \\
\cmidrule{2-3}
& Single Person & Approach the person with a brown shirt and say hi to him \\
\cmidrule{2-3}
& \multirow{3}{*}{Multi Person}  & Approach the person with a grey shirt and say hi to him \\
&              & Approach the person with a brown shirt and say hi to him \\
&              & Approach the person with a black coat and say hi to him \\
\cmidrule{2-3}
& \multirow{3}{*}{Multi Person} & Approach the person with a hat and say hi to him \\
&              & Approach the person with a hat and say hi to him (swap) \\
&              & Approach the person with a mask and say hi to him \\
\cmidrule{2-3}
& \multirow{3}{*}{Multi Person} & Approach the person touching their own head and say hi to him \\
&              & Approach the person with open palm hand and say hi to him \\
&              & Approach the person squatting and say hi to him \\
\cmidrule{2-3}
& \multirow{2}{*}{Multi Person} & Approach the man and say hi to him \\
&              & Approach the woman and say hi to her \\
\cmidrule{2-3}
&\multirow{2}{*}{ Multi Person} & Approach the person on the left and say hi to him \\
&  & Approach the person on the right and say hi to him \\
\midrule
\multirow{3}{*}{Ask the Location}
& \multirow{3}{*}{Single Person} & Approach the person with a black sweater and ask him the location of the classroom \\
&              & Approach the person with a black sweater and ask him the location of the kitchen \\
&              & Approach the person with a black sweater and ask him the location of the restroom \\
\midrule
\multirow{3}{*}{Hand Me Objects}
& \multirow{3}{*}{Single Person} & Approach the person with a grey sweater and ask him to give you some flowers \\
&              & Approach the person with a grey sweater and ask him to give you a cup \\
&              & Approach the person with a grey sweater and ask him to hand you a controller \\
\bottomrule
\end{tabular}}

\end{table*}

\begin{table*}[t]
\centering
\caption{Task prompts for mobile manipulation (pick-and-place) tasks.}
\label{tab:task_prompts}
\resizebox{0.6\linewidth}{!}{
\begin{tabular}{lll}
\toprule
\textbf{Task} & \textbf{Object} & \textbf{Instruction Prompt} \\
\midrule
Pick  & Apple       & Approach and grab the red apple on the desk \\
Pick  & Bottle      & Approach and grab the green bottle on the desk \\
\midrule
Place & Apple       & Approach and place the apple on the desk \\
Place & Bottle      & Approach and place the bottle on the desk \\
\midrule
Pick  & Pen holder  & Approach and grab the black pen holder on the desk \\
Pick  & Pink cup    & Approach and grab the pink cup on the desk \\
\midrule
Place & Pen holder  & Approach and place the black pen holder on the desk \\
Place & Pink cup    & Approach and place the pink cup on the desk \\
\bottomrule
\end{tabular}
}
\end{table*}

\begin{table*}[t]
\centering
\caption{Room navigation and interaction task prompts.}
\label{tab:room_prompts}
\resizebox{0.6\linewidth}{!}{
\begin{tabular}{ll}
\toprule
\textbf{Room} & \textbf{Instruction Prompt} \\
\midrule
\multirow{2}{*}{Room \#1 / \#2 / \#3} & Go into the meeting room \\
                     & Get out of the meeting room \\
\midrule
\multirow{2}{*}{Room \#4}             & Go into the room in front of you \\
                     & Get out of the room \\
\midrule
\multirow{2}{*}{Room \#5}             & Go into the office and say hi to the person \\
                     & Get out of the room \\
\bottomrule
\end{tabular}
}
\end{table*}

\section{Difference between our work and existing work}
\label{app:diff_existing_work}

Our work differs from prior research across three major axes:

\paragraph{Scope of embodiment.} Most VLM-based embodied agents~\citep{NEURIPS2023_efb2072a, wu2023embodied, song2023llm} and LLM-driven systems~\citep{ahn2022can, guo2024embodied, lee-etal-2024-llm} focus on manipulators or simulated agents with simplified embodiments, often relying on predefined skill libraries. In contrast, \projname{} targets full humanoid robots, directly predicting egocentric, low-level actions—including locomotion, posture adjustment, head orientation, manipulation, and human-interaction—bridging abstract instruction reasoning with real-world motor control.

\paragraph{Unified action reasoning.} Existing mobile-manipulation and navigation frameworks~\citep{ehsani2021manipulathor, szot2021habitat, yang2024harmonic} typically decompose tasks into modular subgoals or stage-wise controllers for perception, locomotion, and manipulation. EgoActing and \projname{} unify these components, allowing the model to jointly reason over heterogeneous action types and generate temporally coherent, context-aware sequences without explicit intermediate planning.

\paragraph{Task generalization and real-world deployment.} Vision-and-Language Navigation (VLN)~\citep{anderson2018vision, zhangvision, qi2025vln0r10} and object-goal navigation methods~\citep{chaplot2020object, cao2024cognav0} primarily address static navigation or object-localization tasks. Our approach extends beyond navigation, supporting dynamic, long-horizon tasks that combine movement, manipulation, active perception, and human interaction in cluttered, unseen real-world environments. This enables robust instruction grounding into actionable motor sequences suitable for humanoid robots in practical deployment scenarios.

\end{document}